\documentclass[letterpaper, 10 pt, conference]{ieeeconf}  

\IEEEoverridecommandlockouts                              

\usepackage{graphics, xcolor} 
\usepackage{tabularx}
\usepackage{amsmath, stmaryrd} 
\usepackage{cite}
\usepackage{diagbox}
\usepackage{amsmath,amssymb,amsfonts}
\usepackage{newtxtext,newtxmath}
\usepackage{hyperref}
\usepackage{algorithmic}
\usepackage{graphicx}
\usepackage{textcomp}
\usepackage{xcolor}
\usepackage{balance}
\let\labelindent\relax
\usepackage{enumitem} 
\usepackage{soul}
\makeatletter
\newcommand{\reducedplus}{\mathpalette\reduced@plus\relax}
\newcommand{\reduced@plus}[2]{%
  \sbox6{$\m@th#1+$}%
  \sbox8{\scalebox{0.875}{\copy6}}%
  \dimen@=\dimexpr(\wd6-\wd8)/3\relax
  \raisebox{\dimen@}{\box8}%
}
\newcommand{\boxoperation}[2][\mathbin]{%
  #1{\mathpalette\box@operation{#2}}%
}
\newcommand{\box@operation}[2]{%
  \ooalign{$\m@th#1\boxempty$\cr\hidewidth$\m@th#1#2$\hidewidth\cr}%
}

\def\figref#1{Fig.~\ref{#1}}
\def\tabref#1{Tab.~\ref{#1}}
\def\eqref#1{Eq.~(\ref{#1})}

\usepackage{fancyhdr}
\fancypagestyle{arxivhdr}
{
   \fancyhf{}
   \setlength{\headheight}{15pt} 
\fancyfoot[C]{This paper has been accepted for publication at the 2023 IEEE International Conference on Robotics and Automation (ICRA)\\DOI: 10.1109/ICRA48891.2023.10160758 \url{https://ieeexplore.ieee.org/document/10160758}}
\fancyhead[C]{\footnotesize Please cite this paper as:\\
D. Evangelista, E. Olivastri, D. Allegro, E. Menegatti and A. Pretto, "A Graph-Based Optimization Framework for Hand-Eye Calibration for Multi-Camera Setups," 2023 IEEE International Conference on Robotics and Automation (ICRA), 2023, pp. 11474-11480, doi: 10.1109/ICRA48891.2023.10160758}
}

\title{\LARGE \bf
A Graph-based Optimization Framework for\\Hand-Eye Calibration for Multi-Camera Setups 
}

\author{Daniele Evangelista\textsuperscript{*}, Emilio Olivastri\textsuperscript{*}, Davide Allegro, Emanuele Menegatti and Alberto Pretto%
\thanks{
\textsuperscript{*}These authors contributed equally. All the authors are with the Department of Information Engineering (DEI) at the University of Padova, via Gradenigo 6/B, 35131 Padova, Italy. {\tt\small Emails:[evangelista; olivastrie; allegrodav; emg; alberto.pretto]@dei.unipd.it}.
}}

\begin{document}

\maketitle
\thispagestyle{empty}
\pagestyle{empty}

\thispagestyle{arxivhdr}

\begin{abstract}
Hand-eye calibration is the problem of estimating the spatial transformation between a reference frame, usually the base of a robot arm or its gripper, and the reference frame of one or multiple cameras. Generally, this calibration is solved as a non-linear optimization problem, what instead is rarely done is to exploit the underlying graph structure of the problem itself. 
Actually, the problem of hand-eye calibration can be seen as an instance of the Simultaneous Localization and Mapping (SLAM) problem. Inspired by this fact, in this work we present a pose-graph approach to the hand-eye calibration problem that extends a recent state-of-the-art solution in two different ways: i) by formulating the solution to \textit{eye-on-base} setups with one camera; ii) by covering multi-camera robotic setups. The proposed approach has been validated in simulation against standard hand-eye calibration methods. Moreover, a real application is shown. In both scenarios, the proposed approach overcomes all alternative methods.
We release with this
paper an open-source implementation of our graph-based optimization framework for multi-camera setups. 
\end{abstract}


\section{Introduction}
The ability to robustly perceive the surrounding environment is a key factor for the development and deployment of intelligent robotic work cells in industry. Perception devices, for example cameras, are widely used in industrial robotics applications to perform complex tasks that involves robot arms such as visual inspections \cite{eitzinger_inspection, mapping_etfa} and bin-picking \cite{prettoCASE2013, doi:10.1177/0278364917713117}. They are also involved as assistive devices that continuously monitor the position of human workers in collaborative tasks with robots \cite{hr_collab, human_perception}. However, the information provided by the cameras is generally expressed with respect to their own coordinate systems. 
To make it possible to control a robot arm based on such visual information, it is first necessary to accurately calculate the position of the camera relative to the robot gripper or relative to the robot base or work cell. This process is called \textit{hand-eye calibration}, and this is defined as the procedure to determine the spatial transformation between a robot reference frame (e.g., robot base or end-effector frames) and the reference frame of the camera.
\begin{figure}[t]
    \centering
    \includegraphics[width=\linewidth]{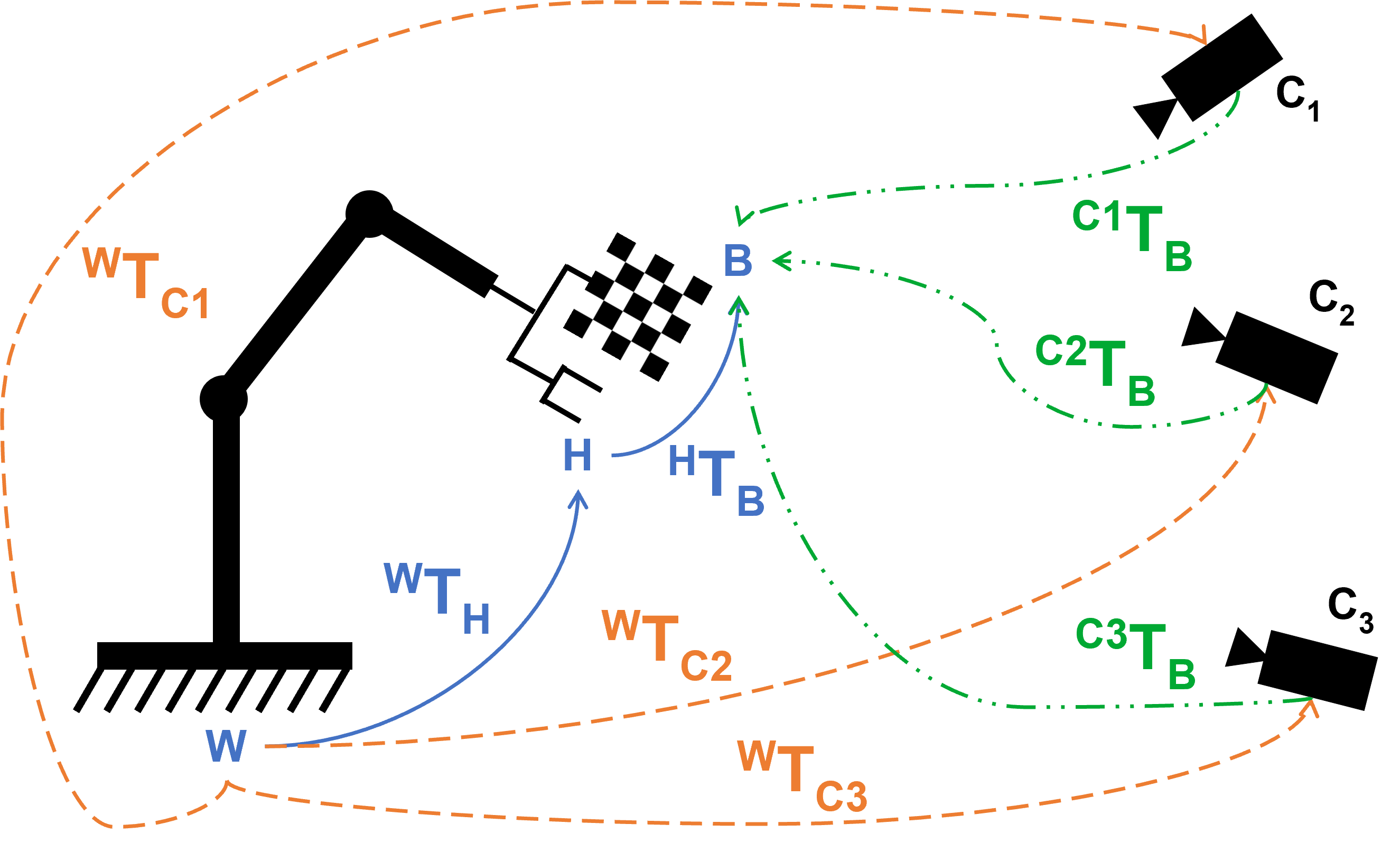}
    \caption{Multi-camera Eye-on-Base calibration setup. The goal of the proposed hand-eye calibration is to compute the position of each camera in the reference frame $W$, i.e., the transformations $^WT_{C1...C3}$ in the figure.}
    \label{fig:camera_setup}
\end{figure}
The hand-eye calibration problem can be classified into two categories\cite{10.1007/s00170-021-08233-6}:
\begin{itemize}
    \item \textbf{Eye-in-hand}: The camera is rigidly mounted on the robot's wrist;
    \item \textbf{Eye-on-base} (or \textbf{Eye-to-hand}): One or more cameras are fixed while the robot arm is moving (e.g., \figref{fig:camera_setup})
\end{itemize}

There exists a variety of methods that solve this problem, many of them based on a global optimization that minimizes specific quantities, such as geometric ones like translation and rotation errors or visual ones like reprojection errors. What, instead, is rarely done is to exploit the underlying graph structure of the problem. In fact, as introduced in \cite{koide}, the hand-eye calibration problem with an eye-in-hand setup can be seen as an instance of the SLAM (Simultaneous Localization And Mapping) problem and can be solved by exploiting standard probabilistic frameworks exploited in SLAM, such as pose graph optimization \cite{5681215}. In this work, we exploit this intuition by proposing a general graph-based tool to solve the complementary eye-on-base calibration problem. To cast this problem to SLAM, one can consider that the motion is relative: instead of considering the motion of the end-effector in front of the static camera, one can imagine the camera is moving, and the robot's end-effector is static. Thus, taking images (i.e., observations) while the robot moves a checkerboard mounted on the end-effector to different poses (i.e., landmarks) results in observations of landmarks used for the optimization.

Our solution is based on the minimization of reprojection error, which is a well-known technique that has been applied to different types of problems (e.g. camera calibration, bundle adjustment, \dots) but surprisingly rarely in the context of hand-eye calibration problems. This technique also allows us to avoid relying strictly on the PnP (Perspective-n-Point) algorithm, which in our case is simply used to infer an initial guess for the optimization.\\

\begin{figure}[t!]
    \centering
    \includegraphics[width=\linewidth]{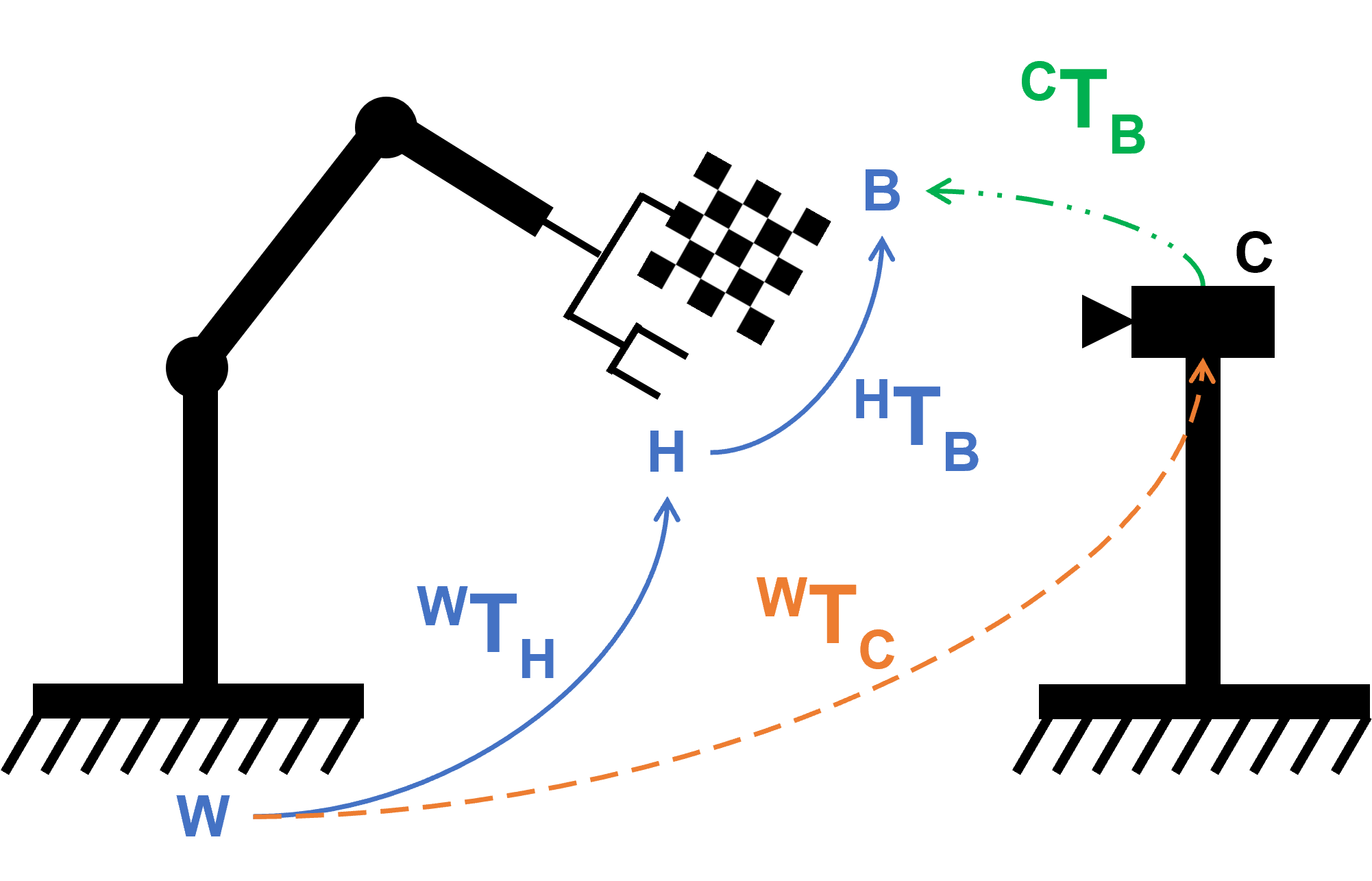}
    \caption{Single-camera Eye-on-Base calibration setup. The goal of hand-eye calibration here is to compute the position of the camera in the $W$ reference frame, namely the transformation $^WT_C$ in the figure.}
    \label{fig:camera_setup_single}
\end{figure}

The main contributions of this work are the following:
\begin{enumerate}[label=(\roman*)]
    \item The proposed approach extends a recent state-of-the-art approach \cite{koide} by implementing a graph-based optimization solution to \textit{eye-on-base} setups with one camera (\figref{fig:camera_setup_single}), not covered in the basic method;
    \item Our method also covers multi-camera setups (\figref{fig:camera_setup}); thus, a general and unified graph-based optimization framework for hand-eye calibration is achieved;
    \item An exhaustive evaluation of the proposed method is reported. The evaluation is primarily performed in simulated environments, where the robustness to visual and geometric noise can be tested. The proposed approach outperforms several popular calibration approaches, proving to be robust to high levels of noise in the calibration data. We also tested our method on a real setup, confirming the convincing results obtained in the simulations;
    \item An open-source implementation of the proposed hand-eye calibration method is made publicly available with this paper at:\\
\url{https://bitbucket.org/freelist/gm_handeye}
    \normalsize
\end{enumerate}
The remainder of this paper is organized as follows. Section~\ref{sec:rel_works} reviews related work on hand-eye calibration. Section \ref{sec:methodology} presents and describes the proposed hand-eye calibration methodology based on the formalization of the process through graph optimization guided by minimization of a customized error function. In Section~\ref{sec:experiments} we show a comprehensive evaluation, performed in simulation, and demonstrate the applicability of the proposed method in real hand-eye calibration problems. Finally, in Section~\ref{sec:conclusions} we draw our conclusions.

\section{Related Work}
In the literature, several methods have been proposed for hand-eye calibration. Shiu and Ahmad \cite{shiu_ahmad} initially proposed a method for the estimation of the hand-eye transformation based on the solution of a homogeneous equation. Tsai and Lenz~\cite{tsai_lorenz} proposed a widely adopted calibration method (e.g., implemented in \cite{visp}) that first estimates the translation and then the rotation of a hand-eye transformation. Similarly to Shiu and Tsai, Chou and Kamel \cite{chou} solved an estimation problem based on a normalized quaternion representation to transform the kinematic equation into two simple and structured linear systems with rank-deficient coefficient matrices. Daniilidis and Bayro-Corrochano~\cite{daniilidis} proposed a similar framework that is based on dual quaternion parameterization and singular value decomposition, and Srobl and Hirzinger~\cite{strobl} introduced a novel metric to optimize the estimation problem by appropriately weighting translation and rotation errors in the optimization problem. Park and Martin \cite{park_handeye} estimated the hand-eye transformation in the Euclidean group using Lie group and least-square optimization.
Gwak \emph{et al.}~\cite{seungwoong} proposed a cyclic coordinate descent algorithm to optimize objective functions in SE(3) and applied it to a class of robotic problems such as hand-eye calibration. In this direction, Horaud and Dornaika~\cite{horaud} proposed a linear formulation of the same optimization problem. The same method has been further applied to online calibration by Andreff \emph{et al.} \cite{andreff}.

More recently Shah \cite{shah_handeye} formulated a closed-form solution for the hand-eye problem by using an SVD-based algorithm and the Kronecker product to solve for rotation and translation separately, while LI  \emph{et al.}~\cite{li_handeye} instead used dual quaternions to solve them simultaneously to overcome the limitations of the Kronecker product.

More general formulations of the hand-eye calibration problem have been done in \cite{single_handeye_icra2021} and \cite{multi_handeye_icra2022}. The latter, in particular, mathematically formulates the problem of camera-to-camera calibration using the geometric constraints of hand-eye calibration when a common reference frame (e.g., robot base frame) is taken into the loop. Although this method formally extends the formulation of \cite{tsai_lorenz} to multi-camera setups, it requires an external motion capture system to accurately recover the position of cameras during calibration, making the whole approach not easily scalable or applicable to small setups.

In our previous work \cite{koide}, we proposed a more general approach based on pose graph optimization that achieves high levels of accuracy and introduces a general solution that can be easily applied to different robotic setups and camera projection models (e.g., both standard pinhole camera projection and X-ray source detector models \cite{evangelista}). Although the proposed method is quite general, it does not cover eye-on-base setups in single-camera setups, and it does not handle multi-camera robotic systems. In this work, we overcome this limitation by extending the approach to eye-on-base setups; moreover, we give a general formulation applicable to multi-camera setups where the position of multiple cameras has to be computed w.r.t. a robot base frame.

\label{sec:rel_works}

\section{Methodology}
\label{sec:methodology}
This section presents a detailed overview of the proposed approach. As already mentioned, this work extends \cite{koide} in two main directions, explained hereafter: eye-on-base for single- and multi-camera setups. In this way, we are able to provide a complete framework that holds the necessary instruments to fulfil all the calibration needs that arise in robotic industrial applications.

\subsection{Eye-on-Base Calibration for Single Camera}


Here, we report the necessary notation for eye-on-base single camera setups.
\begin{itemize}
    \item \textbf{W} : world reference frame, usually placed in the robot's base;
    \item \textbf{H} : hand reference frame, placed in the robot's end-effector;
    \item \textbf{B} : checkerboard reference frame;
    \item \textbf{C} : camera reference frame;
\end{itemize}
While the transformations are :
\begin{itemize}
    \item $\bf ^{W}T_{H}$ : Isometry representing the pose of the end-effector in the world reference frame;
    \item $\bf ^{H}T_{B}$ : Isometry representing the pose of the checkerboard in the end-effector reference frame;
    \item $\bf ^{C}T_{B}$ : Isometry representing the pose of the checkerboard in the camera reference frame;
    \item $\bf ^{W}T_{C}$ : Isometry representing the pose of the camera in the world reference frame.
\end{itemize}

\begin{figure}[t!]
    \centering
    \includegraphics[width=\linewidth]{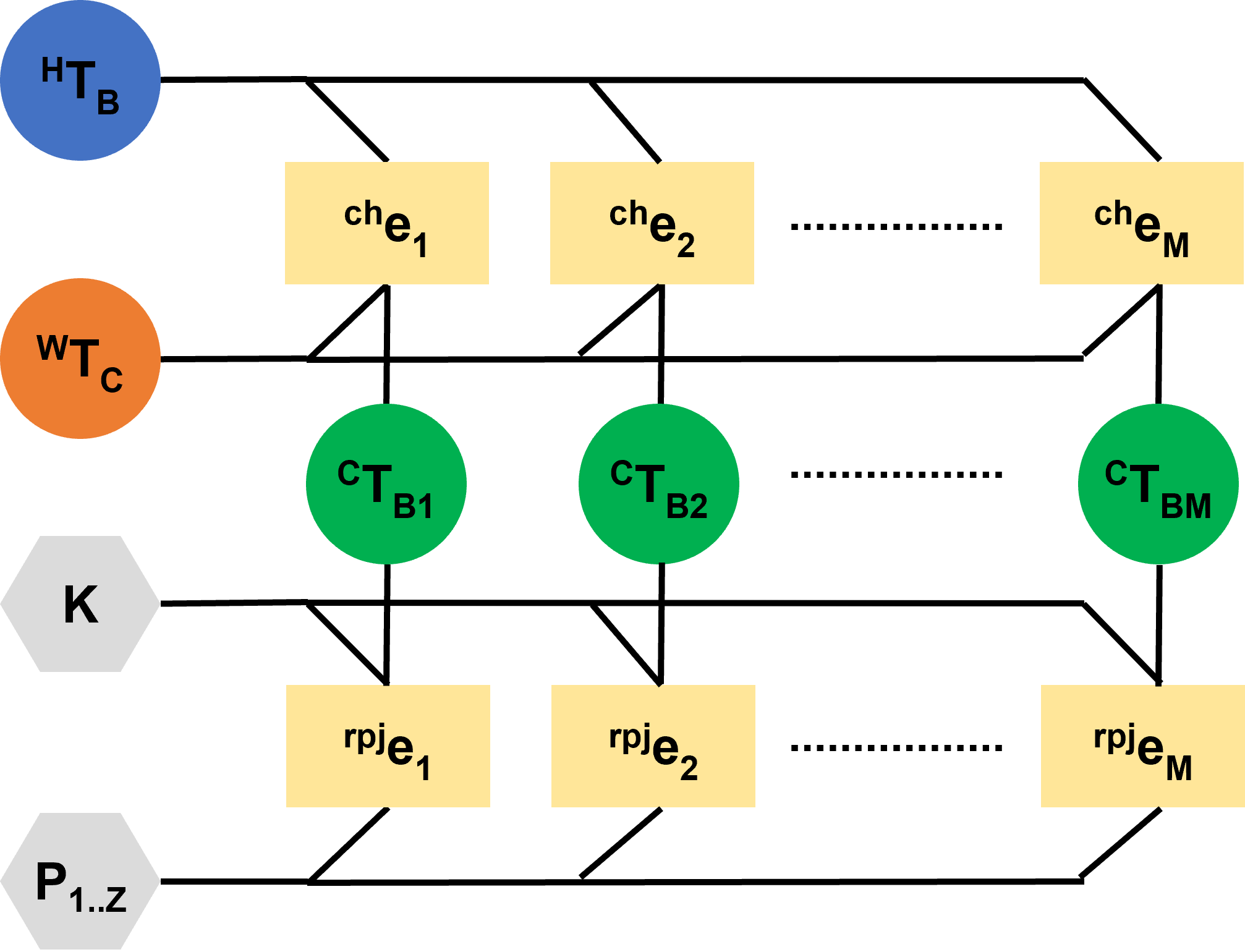}
    \caption{Graphical representation of the optimization problem for eye-on-base single camera setups. The representation contains circles, rectangles and hexagons that respectively represent the estimated variables, error functions, and fixed parameters. The error functions are defined following the equations \ref{eq:long_chain_error} and \ref{eq:rpj_error}.}
    \label{fig:single_graph}
\end{figure}

\figref{fig:single_graph} shows the graph structure for the eye-on-base calibration. Using the notation from
\cite{koide}: circles represent the state variables that need to be estimated, while hexagons are fixed parameters such as $K$ and $P_{1..Z}$, that represent the camera matrix and the set of corners points of the checkerboard in their reference system, respectively. $M$ represents the number of observations, i.e., images and robot poses, used for the calibration. The rectangles represent the different error terms. The error is generally defined as 

\begin{equation}
    \label{eq:general_error}
   e_{i}(x) = h_{i}(x) \boxminus z_i,
\end{equation}

where $h_{i}(x)$ is the measurement function and $z_i$ is the i-th observation and $\boxminus$ is the difference operator for the manifold domain.\\
The first type of edge that is going to be presented is the one that enforces the equality :
\begin{equation}
    \label{eq:long_chain}
   \bf {^W}T_{H} \cdot {^H}T_{B} = {^W}T_{C} \cdot {^C}T_{B},
\end{equation}
that is one of the classic constraints for eye-on-base calibration \cite{axzb}. And its corresponding error representation:
\begin{equation}
    \label{eq:long_chain_error}
   ^{ch}\mathbf{e}_i = t2v([\mathbf{{{^H}T_{W}}]_i \cdot {^W}T_{C} \cdot [{^C}T_{B}]_i \cdot {^B}T_{H}}),
\end{equation}

where the observation $ \mathbf{z}_i = [\bf ^{W}T_{H}]^{-1}_i$ is the $i$-th pose of the end-effector in the world. The remaining transformations are the ones that need to be estimated, while the $t2v( \cdot )$ is a function that converts the transformation to the corresponding minimal manifold representation in order to prevent optimization degradation due to over parametrization.
The second type of edge instead enforces the reprojection error constraint: 
\begin{equation}
    \label{eq:rpj_error}
   ^{rpj}\mathbf{e}_i = \begin{bmatrix}
                \pi (\bf [{^C}T_{B}]_i \cdot {^B}P_1) - \bf [u_{1}]_i\\
                \vdots \\
                \pi (\bf [{^C}T_{B}]_i \cdot {^B}P_j) - \bf [u_{j}]_i \\
                \vdots \\
                \pi (\bf [{^C}T_{B}]_i \cdot {^B}P_Z) - \bf [u_{Z}]_i
                \end{bmatrix},
\end{equation}
where the observation $ \mathbf{z}_i = \bf [u_1 \dots u_j \dots u_z ]_i$ is a vector of tuples $ \mathbf{u_j} = (u_x, u_y)_j $ that are the pixel coordinates of the $j$-th detected corner of the checkerboard at the $i$-th observation and $\pi$ is the projection function of the camera.\\
The optimization aims to minimize the following function:
\begin{equation}
    \label{eq:simple_target_function}
   \min\sum_{i=1}^{M}\left( \lambda \| ^{rpj}\mathbf{e}_i \|^2 + \| ^{ch}\mathbf{e}_i \|^2 \right),
\end{equation}
where $\lambda$ is a weighting factor that accounts for the non homogeneity of the two error terms.
The minimization is carried out by using the Levenberg-Marquardt \cite{more1978levenberg} algorithm.
Our method proved to be robust also if provided with an identity transformation as initial guess, showing a rather large basin of convergence. This means that the system can be correctly calibrated even if we do not provide a good initial guess, which could be difficult to compute.

\subsection{Eye-on-Base Calibration for Multi-Camera}
\begin{figure}[t]
    \centering
    \includegraphics[width=\linewidth]{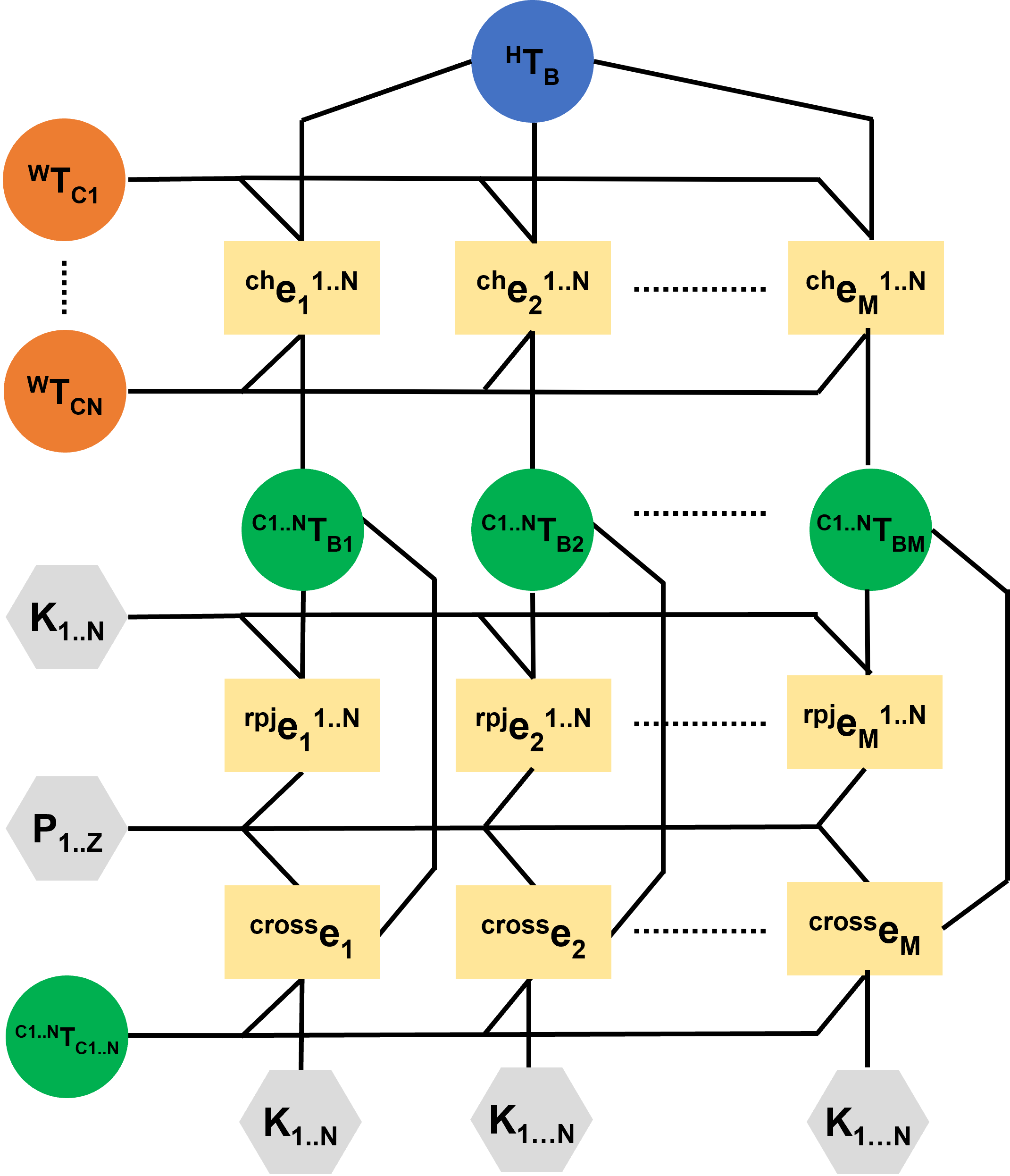}
    \caption{Evolution of the graph representation in multi cameras setups. The figure reports the error terms of Eq. (\ref{eq:stack_eq_err}-\ref{eq:stack_rpj_err}) that are the stack of the errors in Eq. (\ref{eq:long_chain_error}-\ref{eq:rpj_error}), respectively. Moreover, the graph includes an additional error defined in Eq. \ref{eq:cross_err_func} which enforces constraints between the cameras with overlapping field of view.}
    \label{fig:multi_graph}
\end{figure}
If the robot is observed by multiple static cameras, one could repeat the above algorithm several times. On the other hand, running an algorithm that performs single-camera calibration $N$ times, with $N$ the number of cameras that are present in the workspace, is a sub-optimal solution because, in the optimization, the cross-observations between the different cameras are not factorized.\\
Given $N$ cameras in the workspace, as shown in \figref{fig:camera_setup}, there is the need to introduce a little more notation:
\begin{itemize}
    \item $\bf {{^{C_d}}}T_{C_a}$ : Isometry representing the pose of the $a$-th camera in the $d$-th camera reference frame;
    \item $\bf {^{C_a}}T_{B}$ : Isometry representing the pose of the checkerboard in the $a$-th camera reference frame;
    \item $\bf {^{W}}T_{C_a}$ : Isometry representing the pose of the $a$-th camera in the world reference frame.
\end{itemize}
The starting point for the multi-camera model is simply stacking the single-camera model $N$ times as the number of the cameras.
In \figref{fig:multi_graph} some of the vertices and edges have been grouped together in order to facilitate the graphical representation.\\
The error term $^{ch}\mathbf{e}_i^{1..N} $ enforces all the chained equalities of the system :
\begin{equation}
    \label{eq:stack_eq_err}
    ^{ch}\mathbf{e}^{1..N}_i = \begin{bmatrix}
                    ^{ch}\mathbf{e}_i^{1} \ldots ^{ch}\mathbf{e}_i^{a} \ldots ^{ch}\mathbf{e}_i^{N}
                    \end{bmatrix}^T,
\end{equation}
where $^{ch}\mathbf{e}_i^{a}$ represents the chained equality for the $a$-th camera at the $i$-th time step.

The error term relative to the reprojection error evolves into:
\begin{equation}
    \label{eq:stack_rpj_err}
   ^{rpj}\mathbf{e}^{1..N}_i = \begin{bmatrix}
                      ^{rpj}\mathbf{e}_i^{1} \ldots ^{rpj}\mathbf{e}_i^{a} \ldots ^{rpj}\mathbf{e}_i^{N}
                      \end{bmatrix}^T,
\end{equation}
where $^{rpj}\mathbf{e}_i^{a}$ is the reprojection error term relative to the camera $a$-th at the $i$-th time step.
Up to this point, we have simply stacked the single-camera graph $N$ times one on top of the other, in fact the derivation of the new error terms was straightforward. The additional error term that is added to the graph model is the one that represents the cross-observation between cameras, from which information about their relative spatial relation can be extracted.
Cross-observations occur when more cameras are able to see the checkerboard at the same time step $i$. In order to add it in an automatic manner to the optimization, it is needed to build a co-visibility matrix $\mathbf{X}_i$. At each time step $i$, each cell of this matrix is represented by $\mathbf{X}_i(a, d)$, where $a$ and $d$ are two different cameras (i.e., $ a\neq d $). The cell contains a binary value that encodes whether or not the checkerboard is in the field of view of both cameras $a$ and $d$ at the $i$-th time step:
\begin{equation}
    \label{eq:selective_variable}
    \mathbf{X}_i(a,d)=
    \begin{cases}
      0, & \text{if}\ a=d \text{ or no cross-observation} \\
      1, & \text{otherwise}
    \end{cases}
\end{equation}

Thus, using the co-visibility matrix we are able to enforce the cross-observation constraints only where they are actually aiding the optimization.

To capture the spatial relation between cameras the transformation $\bf ^{C_d}T_{C_a} $ is added as a state variable that needs to be estimated. 

The new error term that models the cross-observation is defined as follows:
\begin{equation}
\label{eq:cross_rpj_err}
^{cross}\mathbf{e}_{iad} = \begin{bmatrix}
                  \pi_a(\mathbf{^{C_a}T_{C_d} \cdot [^{C_d}T_{B}]_i \cdot P_1^B}) - \bf [u_1^a]_i\\
                  \vdots \\
                  \pi_a(\mathbf{^{C_a}T_{C_d} \cdot [^{C_d}T_{B}]_i \cdot P_j^B}) - \bf [u_{j}^a]_i\\
                  \vdots \\
                  \pi_a(\mathbf{^{C_a}T_{C_d} \cdot [^{C_d}T_{B}]_i \cdot P_Z^B}) - \bf [u_Z^a]_i
                  \end{bmatrix},    
\end{equation}
where the indices $a$ and $d$ refer to the cameras and $\bf [u_{j}^a]_i$ represents the $j$-th corner of the checkerboard detected by the camera $a$ at the $i$-th time step.
This error accounts for the current estimate of the relative transformation between cameras. Checkerboard points are projected into the camera $c_a$ using the camera $c_d$ as starting point.\\

To simplify the notation all the error terms at time step $i$ will be marginalized:
\begin{equation}
    \label{eq:cross_err_func}
    ^{cross}\mathbf{e}_{i} = \sum_{a=1}^{N} \sum_{d=1}^{N} \| ^{cross}\mathbf{e}_{iad} \|^2 \cdot \mathbf{X}_i(a,d), \text{with } a \neq d.
\end{equation}

\figref{fig:multi_graph} shows the complete graphical model that includes these final constraints. The resulting function that needs to be minimized is the following:
\begin{equation}
    \label{eq:target_function}
    \min\sum_{i=1}^{M}\left( \lambda_1 \| ^{rpj}\mathbf{e}_i \|^2 + \| ^{ch}\mathbf{e}_i \|^2 +\  \lambda_2^{cross}\mathbf{e}_{i} \right)
\end{equation}
The effect that the cross-observations have on the optimization is to make it more robust against noise, and it will be effectively shown in the following section.

The proposed algorithm has been implemented using the g2o \cite{kummerle2011g} framework. The weighting factors $\lambda_1$ and $\lambda_2$ have been experimentally tuned and set to the values $[10^{-6}, 10^{-3}]$, respectively, and kept fixed during all experiments.

\begin{figure*}[t!]
    \centering
    \includegraphics[width=0.95\linewidth]{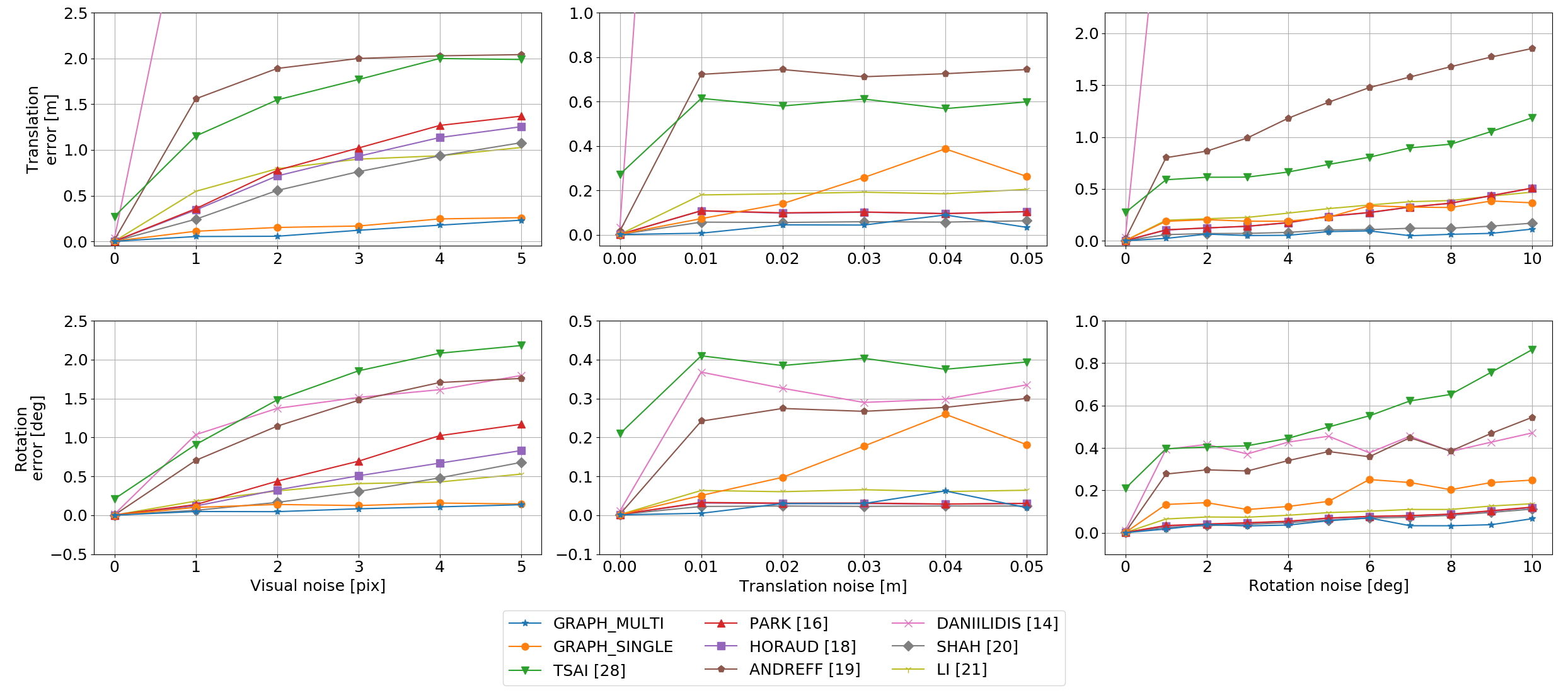}
    \caption{Result of the simulated evaluation. The proposed approach, both in \textit{single-} and \textit{multi-camera} settings has been tested against Tsai \cite{tsai_lenz}, Park \cite{park_handeye}, Horaud \cite{horaud}, Andreff \cite{andreff}, Daniilidis \cite{daniilidis}, Shah \cite{shah_handeye} and Li \cite{li_handeye}.}
    \label{fig:all_methods_experiments}
\end{figure*}

\begin{figure*}[t!]
    \centering
    \includegraphics[width=0.95\linewidth]{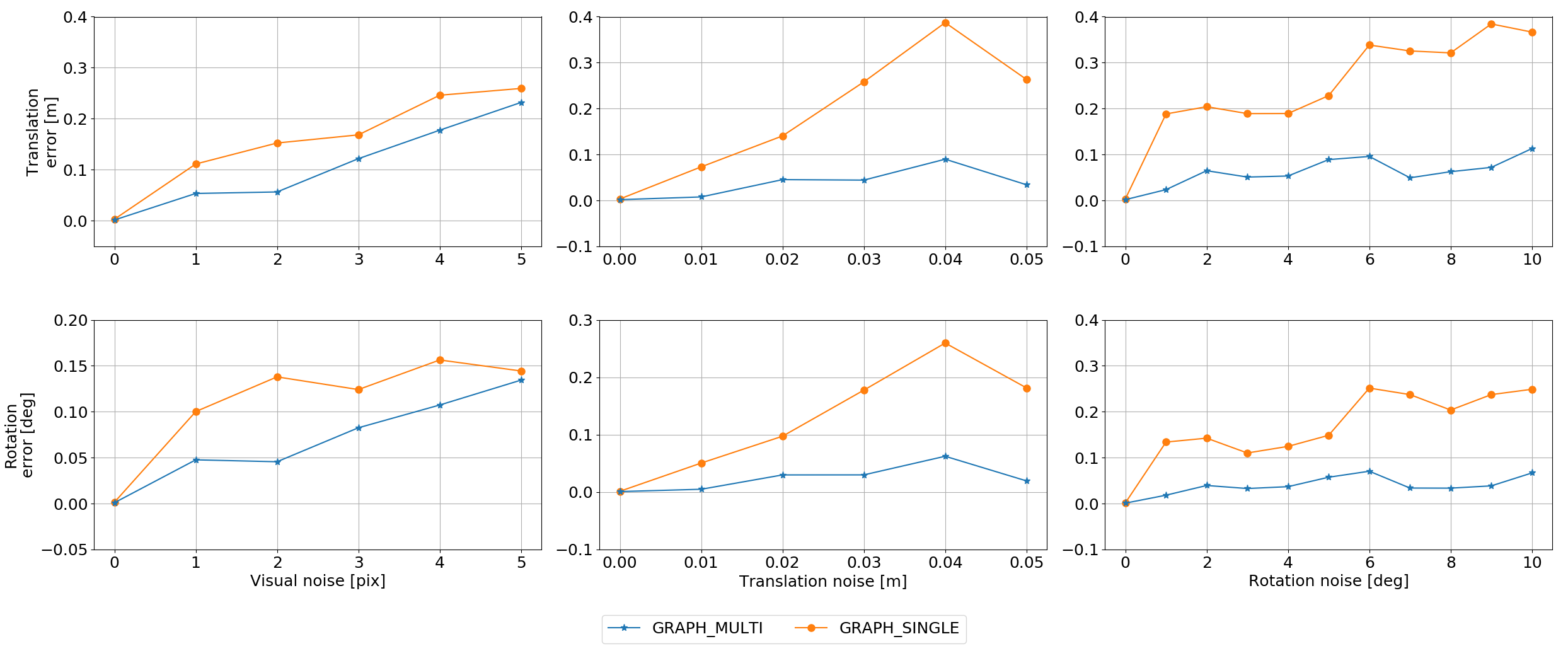}
    \caption{More in depth focus of the simulated evaluation regarding the proposed method only in \textit{single-} and \textit{multi-camera} settings.}
    \label{fig:only_ours_experiments}
\end{figure*}

\begin{table*}[t!]
    \begin{center}
    \resizebox{\linewidth}{!}{
        \begin{tabular}{|c||c||c||c||c||c||c||c||c|}
             \hline
              \diagbox[width=10em]{Camera}{Error\\$[pix]$}  &  TSAI \cite{tsai_lenz} & PARK \cite{park_handeye} & HORAUD \cite{horaud} & ANDREFF \cite{andreff} & DANIILIDIS \cite{daniilidis} & SHAH \cite{shah_handeye} & LI \cite{li_handeye} & GRAPH\_MULTI \\
             \hline
             Camera 1  & 443.763  & 0.520446 & 0.432767 & 1.40595 & 196253 & 0.503051 & 0.603978 &\textbf{0.337803}  \\
            \hline
             Camera 2  & 0.416451  & 0.346938 & 0.35017 & 1.98045 & 0.538384 & 0.439182 & 0.717876 &\textbf{0.284408}  \\
            \hline
             Camera 3  & 148.89  & 26.2883 & 25.0635 & 165429 & 38.9004 & 97.8753 & 417.908 &\textbf{4.4232}  \\
            \hline
             Average  & 197.69 & 9.0519 & 8.61548 & 55144.2 & 65430.8 & 32.9392 & 139.743 &\textbf{1.6818}   \\
            \hline
            \end{tabular}
            }
        \caption{Evaluation of the reprojection errors using the real setup data. All the values in the table are in pixel unit.}
        \label{tab:reaL_table}
    \end{center}
\end{table*}

\section{Experiments}
\label{sec:experiments}
In Section~\ref{subsec:simulation} we show the robustness and estimation accuracy of the proposed method by testing it under visual and geometric perturbations. In particular, we both added noise to the detection of the checkerboard corners, thus simulating wrong detections of the calibration pattern (e.g., when using low-resolution cameras), and we also added noise to the robot poses $\bf [{^H}T_{W}]_i$ used for calibrating. 

The proposed approach has been tested against the following methods (i.e., implementation from OpenCV\footnote{\url{http://docs.opencv.org/4.5.4/d9/d0c/group__calib3d.html}}): \cite{tsai_lenz, park_handeye, horaud, andreff, daniilidis, shah_handeye} and \cite{li_handeye}. In the graphs and in the table we name them by reporting the first author of each one. The alternative methods have been tested by calibrating every single camera independently and then averaging the errors. This experimental methodology has been adopted because the alternative approaches are not directly usable for multi-camera calibration. The evaluation also reports an in-depth focus on the comparison of our multi-camera calibration method against the repetition of the proposed single-camera calibration applied to the individual cameras, this is done to demonstrate the robustness and improved accuracy of the multi-camera calibration framework.

\subsection{Experiments with Synthetic Data}
\label{subsec:simulation}
Using the same evaluation protocol proposed in \cite{koide, evangelista_etfa2022}, our method has been tested, under simulated conditions, using three different sources of noise: visual, translation, and rotation. This evaluation protocol aims to test the robustness of the calibration under perturbed visual conditions (i.e., introducing noise in the detection of the corners of the calibration pattern), and it is also used to test the optimization accuracy and convergence capabilities when additional geometric noise is added to the robot poses.

We collected simulated data using the \emph{Gazebo}\footnote{\url{https://gazebosim.org/home}} simulator, in which a simulated work cell has been implemented. The proposed work cell is composed of an industrial manipulator, mounting the calibration pattern, surrounded by five external cameras as in typical multi-camera eye-on-base setups. A set of 150 robot poses has been selected by randomly sampling the robot workspace to uniformly cover each camera field of view.

The results of this evaluation are reported in \figref{fig:all_methods_experiments} and \figref{fig:only_ours_experiments}. In particular, in \figref{fig:all_methods_experiments} the plots show the performance of all the calibration methods, while in \figref{fig:only_ours_experiments} only the results of our single- and multi-camera graph-based calibration approaches are reported. The results show how our proposed method achieves good results in terms of translation and rotational errors even in strong perturbance conditions, demonstrating that it is more robust compared to other methods. The proposed eye-on-base single-camera calibration (i.e., GRAPH\_SINGLE in \figref{fig:all_methods_experiments} and \figref{fig:only_ours_experiments}) generally outperforms all the other tested approaches. Moreover, our multi-camera calibration hand-eye algorithm (i.e. GRAPH\_MULTI in \figref{fig:all_methods_experiments} and \figref{fig:only_ours_experiments}) is the best performing approach, in particular when high levels of noise are present in the calibration data.

\begin{figure}[t!]
    \centering
    \includegraphics[width=\linewidth]{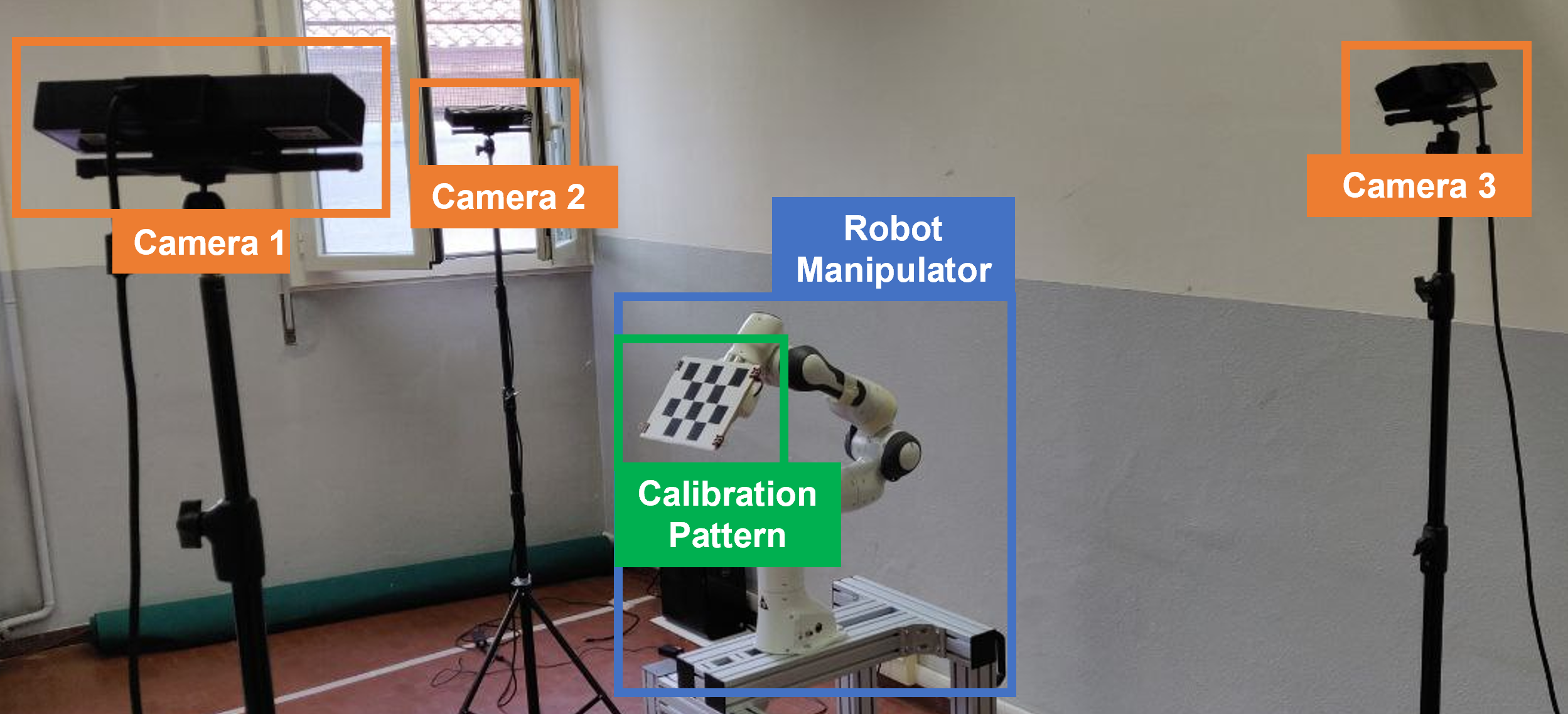}
    \caption{Front view of the real setup where the proposed calibration method has been tested.}
    \label{fig:real_front}
\end{figure}

\subsection{Experiments on a Real Setup}
\label{subsec:real}
\begin{figure}[t!]
    \centering
    \includegraphics[width=\linewidth]{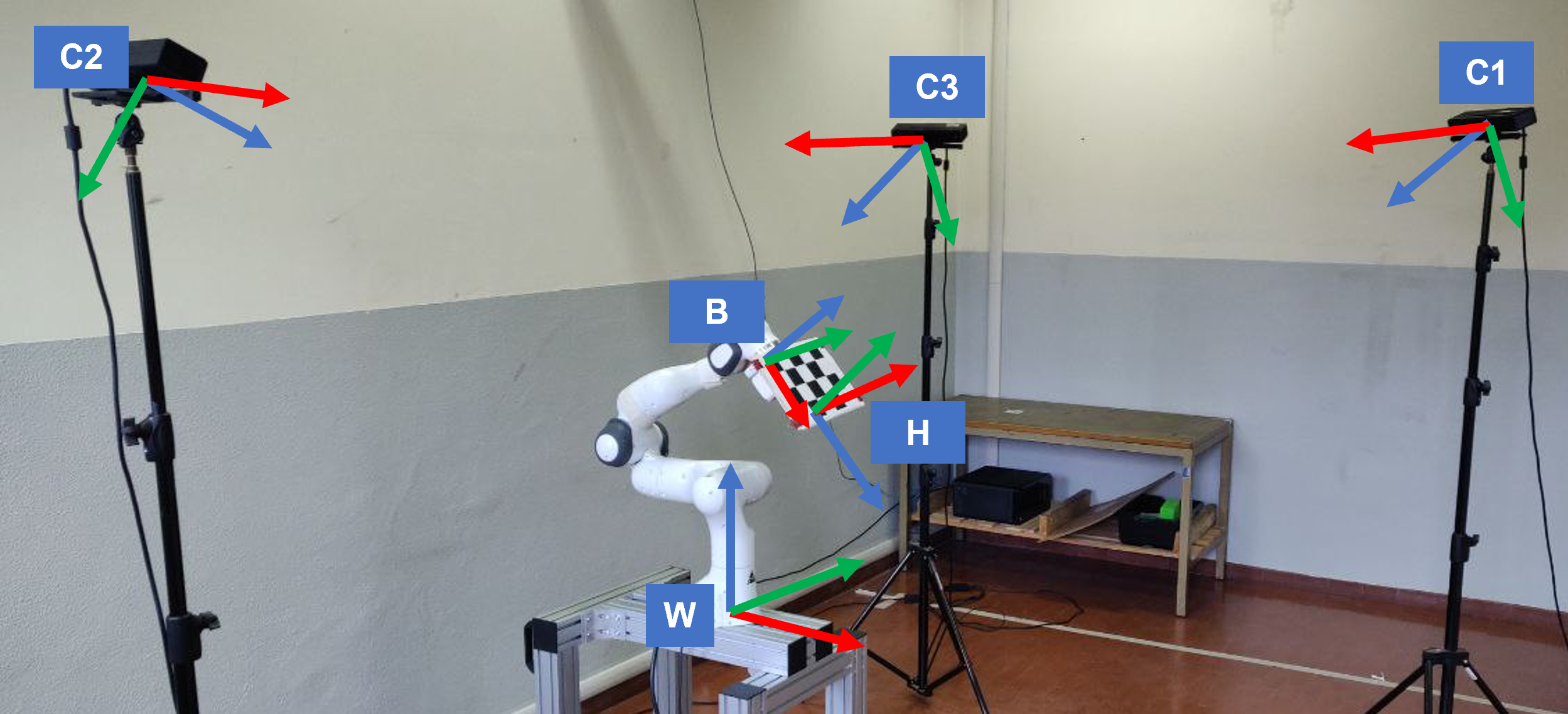}
    \caption{Rear view of the real setup. Main reference frames are depicted.}
    \label{fig:real_rear}
\end{figure}

The proposed algorithm has also been validated in a real robotic environment. The setup consisted of a Franka Emika Panda\footnote{\url{https://www.franka.de/}} robot manipulator and three Kinect2 cameras (we use only the RGB images) placed around the robot, as shown in \figref{fig:real_front} and \figref{fig:real_rear}. Unlike the simulated evaluation, where ground truth is available, in the real environment the absolute pose of each camera is not known a priori. Thus, to perform the test in the real setup, the reprojection error has been considered as a metric. In \tabref{tab:reaL_table} the results of all approaches are reported. 


Our approach is capable of correctly calibrating all cameras in the setup, also in challenging lighting condition, e.g., for cameras \#1 and \#3 (see \figref{fig:real_front}). Most of the tested methods were unable to converge to a low reprojection error solution, while our method provided the lowest error by a large margin.

\section{Conclusions}
\label{sec:conclusions}
This work presents a pose-graph solution to the hand-eye calibration problem that extends previous state-of-the-art work \cite{koide} in two different ways: i) by formulating the solution to \textit{eye-on-base} setups with one camera; ii) by extending the  \textit{eye-on-base} calibration to multi-camera setups. To validate the proposed system, we used the same experimental protocol proposed in \cite{koide} and \cite{evangelista_etfa2022}. 

The proposed approach has shown to be robust when the calibration data are affected by visual and geometric errors. Compared to traditional solutions, the implemented pose-graph optimization achieves better accuracy in terms of geometric errors (i.e., translation and rotation). 

The proposed approach also proves to be robust and accurate when tested in real hand-eye calibration benchmarks, where absolute ground truth is not available and accuracy is measured in terms of reprojection error.

It should be noticed that the proposed approach, as in the other hand-eye calibration methods used for the experiments, does not take into account the influence of dynamic effects given by gravity or joint and link flexibility. This may affect the accuracy and precision of robot manipulators and must be taken into account with proper dynamic calibration techniques. 

The authors believe that this kind of calibration can be formulated with the same graph-based structure as in the proposed hand-eye calibration approach, and it may be considered as future work in this field to reach a more general calibration procedure. Moreover, the proposed graph-based formulation can be further extended in future research to address more complex problems such as calibration of \emph{multi-robot-multi-camera} setups. 

\pagebreak

\balance
{
\bibliographystyle{IEEEtran}
\bibliography{references}
}

\end{document}